\documentclass[pdflatex,sn-mathphys]{sn-jnl}% Math and Physical Sciences Reference Style
\usepackage{textcomp}
\usepackage{adjustbox}
\usepackage{rotating}
\usepackage{expex}

\jyear{2022}%

\theoremstyle{thmstyleone}%
\theoremstyle{thmstyletwo}%

\theoremstyle{thmstylethree}%
\newcommand{\suzan}[1]{\textcolor{black}{#1}}
\newcommand{\revision}[1]{\textcolor{black}{#1}}

\begin{document}

\title[Political corpus creation through automatic speech recognition on EU debates]{Political corpus creation through automatic speech recognition on EU debates}


\author[1]{\fnm{Hugo} \spfx{de} \sur{Vos}}\email{h.p.de.vos@fgga.leidenuniv.nl}

\author*[2]{\fnm{Suzan} \sur{Verberne}}\email{s.verberne@liacs.leidenuniv.nl}


\affil[1]{\orgdiv{Department of Public Administration}, \orgname{Leiden University}, %\orgaddress{\street{Street}, 
\city{The Hague}, %\postcode{100190}, \state{State}, 
\country{The Netherlands}
}

\affil*[2]{\orgdiv{Leiden Institute of Advanced Computer Science}, \orgname{Leiden University}. %\orgaddress{\street{Street}, 
\city{Leiden}, %\postcode{10587}, \state{State}, 
\country{The Netherlands}
}

%\affil[3]{\orgdiv{Department}, \orgname{Organization}, \orgaddress{\street{Street}, \city{City}, \postcode{610101}, \state{State}, \country{Country}}}


\abstract{%The abstract serves both as a general introduction to the topic and as a brief, non-technical summary of the main results and their implications. Authors are advised to check the author instructions for the journal they are submitting to for word limits and if structural elements like subheadings, citations, or equations are permitted.
%1. what did we do
In this paper, we present \revision{a transcribed corpus} of the LIBE committee of the EU parliament, totalling 3.6 Million running words.
%2. why did we do it
The meetings of parliamentary committees of the EU are a potentially valuable source of information for political scientists but the data is not readily available because only disclosed as speech recordings together with limited metadata. \revision{The meetings are in English, partly spoken by non-native speakers, and partly spoken by interpreters.} \revision{We investigated the most appropriate Automatic Speech Recognition (ASR) model to create an accurate text transcription of the audio recordings of the meetings in order to make their content available for research and analysis.}
%3. how did we do it
We focused on the \revision{unsupervised} domain adaptation of the ASR pipeline. Building on the transformer-based Wav2vec2.0 model, we experimented with multiple acoustic models, language models and the addition of domain-specific terms. 
%4. what did we find
We found that a domain-specific acoustic model and a domain-specific language model give substantial improvements to the ASR output, reducing the word error rate (WER) from 28.22 to 17.95. The use of domain-specific terms in the decoding stage did not have a positive effect on the quality of the ASR in terms of WER. Initial topic modelling results indicated that the corpus is useful for downstream analysis tasks.
%5. what do we think it means
We release the resulting corpus and our analysis pipeline for future research.\footnote{All our code and data can be found at \\\url{https://github.com/hdvos/EUParliamentASRDataAndCode}}
}




\maketitle

\section{Introduction}\label{sec:Intro}

The transcripts of the plenary sessions of the European Parliament are often used for political \revision{\cite{beaton2007interpreted,greene2015unveiling} as well as linguistic research~\cite{monti2005studying,garssen2017role}}. This data has many advantages, of which the most prominent is accessibility: all transcripts are available on the website of the European Parliament and almost all the transcripts are translated into all official languages of the EU. 
Together with the written transcripts, there is also the possibility to listen \revision{to} and view the meetings, as audiovisual materials are published on the website of the EU. 
In prior work, EU-parliament text corpora have been released for computational purposes such as in the Europarl corpus \cite{koehn_europarl_nodate}.

There are, however, also several drawbacks to the plenary session transcripts as research data for political scientists. 
While they are of good value for determining general trends and topics that are discussed in the EU, content-wise they are limited: Speaking times \revision{for Members of the European Parliament (MEPs)} in the plenary sessions are generally limited to one minute\footnote{\url{https://www.europarl.europa.eu/doceo/document/RULES-9-2022-07-11-RULE-194_EN.html}}, and, as a result, the content of the speeches is typically concise and shallow and therefore not a good representation of the depth of the debate.

In addition to the plenary sessions, the EU parliament comprises of \revision{30} committees\footnote{as of \revision{February 2023} according to \url{https://www.europarl.europa.eu/committees/en/about/list-of-committees}} concerning different topics.
In these committees, topics are discussed in more detail than in the plenary sessions. \revision{The recordings of these meetings, therefore, are a potential goldmine of information: While in the plenary meetings, MEPs give short explanations on their votes for not longer than one minute, in the committee meetings the actual political debate takes place. These committee meetings shed more light on political disagreements between members and how parties, as well MEPs, deal with politically complex issues. In terms of discourse style, the discourse in the committee meetings resembles spontaneous speech more than the discourse in the plenary meetings does. }
If political scholars want to understand how MEPs or party groupings assess specific proposals and determine their position, they need to have access to the content of these committee meetings. \revision{In addition, the data could be a valuable source to linguists interested in political discourse or the role of interpreters.} %Reviewer: (especially as available parliamentary data, such as official Hansards, tends to be edited to remove many spoken features).

Despite a potentially rich source of information, the use of those committee meetings for political research is unfortunately limited because there are no transcripts of those meetings. 
For each meeting, only an agenda, minutes, and an audiovisual recording are available. 
The agenda and minutes are concise and do not contain detailed information on the contents and viewpoints of the different parties. The problem of the recordings is that it is labor intensive to study them \revision{in their original form:} Studying one meeting requires listening to a 3 hour long recording and the average number of meetings per year for one committee is $66.4$.

For these reasons, we \revision{investigate the most appropriate Automatic Speech Recognition (ASR) model to create an accurate text transcription of the audio recordings of the meetings in order to make their content available for research and analysis.} In particular, we employ the Wav2Vec2.0 algorithm \cite{baevski2020wav2vec} to automatically transcribe all meetings of the Committee on Civil Liberties, Justice and Home Affairs (LIBE committee) during the EU-parliamentary sessions of 2014-2019. 

\revision{The language spoken in the meetings is English, in two different variants: around half of the utterances is in English spoken by participants of the meeting, who often are non-native speakers of English; the other half is the English spoken by an interpreter, who in real time translates the utterance of a non-English speaker to English. In fragments with both a non-English speaker and an interpreter, the interpreter language is dominant in the recording but it is preceded by a short fragment of non-English spoken before the interpreter starts translating.}

\revision{The non-native English as well as the combination of non-English and interpreter English require a robust ASR model. In addition, we expect the domain-specific language of the political discussions to require domain-specific ASR models. Therefore, in the ASR pipeline}, we experiment with two different Wav2Vec2.0 models and with the use of a domain-specific language model to decode the output. \revision{We address the following research questions:} %For the latter, we optimized the parameters. 
\begin{enumerate} 
    \item How much improvement can be gained by replacing the pre-trained acoustic model by a domain-specific model for parliamentary data?
    \item How much improvement can be gained by the addition of a domain-specific language model as decoder of the ASR model? 
    \item What is the effect of boosting domain- and context-specific terms extracted from the meeting metadata in the ASR?
\end{enumerate}

Our contributions are twofold: %(1) we show that \textcolor{red}{the state of the art in ASR can be used to generate large scale text corpora;} 
(1) We show that using a domain-specific acoustic model and an adapted language model substantially improves the quality of transformer-based speech recognition in the political domain; 
(2) we deliver a corpus of transcribed LIBE meetings as well as a pipeline that can be used to transcribe EU parliament meetings in English. % (3) we find that \textcolor{red}{the quality of the automatic transcriptions is (and never will be) perfect but is sufficient for a lot types of political research.}}

\section{Background}\label{sec:related_work}

% https://scholar.google.nl/scholar?hl=en&as_sdt=0%2C5&q=transcripts+of+the+plenary+sessions+of+the+European+Parliament+&btnG=

% The IBM 2007 speech transcription system for European parliamentary speeches. B Ramabhadran, O Siohan… - 2007 IEEE Workshop on …, 2007 - ieeexplore.ieee.org

% An approach to corpus-based interpreting studies: developing EPIC (European Parliament Interpreting Corpus) C Bendazzoli, A Sandrelli - Proceedings of Challenges of …, 2005 - academia.edu

%The debates of the European Parliament as linked open data A Van Aggelen, L Hollink, M Kemman… - Semantic …, 2017 - content.iospress.com

% The isl 2007 english speech transcription system for european parliament speeches S Stüker, C Fügen, F Kraft, M Wölfel - Eighth Annual Conference of …, 2007 - academia.edu

In prior work addressing ASR in the political domain \cite{deVos2020challenges}, we found that in general the quality of out-of-the-box ASR looked promising, especially since most of the speech was non-native, but the main drawback was the recognition of names and institutions, as well as some domain specific terms such as ``visa''. 
That motivated the use of open-source ASR methods that allows adaptation and fine-tuning towards the political domain.

\subsection{Wav2Vec2.0}\label{subsec:backgr:wav2vec}

Wav2vec2.0~\cite{baevski2020wav2vec} is a method for self-supervised learning of speech representations with transformers. \revision{Self-supervised learning is a form of supervised learning without the addition of explicit labels to the data; aspects of the data itself are used as labels on large amounts of data. For text data, the commonly used self-supervised learning paradigm is masked language modelling~\cite{devlin2019bert}: masking words and predicting the best fitting word for the context. For speech data, Baevski et al.~\cite{baevski2020wav2vec} learn representations of audio by pre-training on the task of identifying the true audio fragment for a masked time step among a set of candidate audio fragments. } %It works equivalently to how BERT learns word representations in a self-supervised manner. 
%The biggest challenge in the case of speech is to turn a continuous speech signal into discrete units so that a transformer model can be applied. For this Baevski et al.~\cite{baevski2020wav2vec} use a technique called Product Quantization which is described in more detail in \cite{baevski2019vq} and \cite{jegou2010product}

\revision{With supervision of human-transcribed data, a} wav2vec2.0 model can be fine-tuned towards specific (speech recognition) tasks, for example in low resource languages \cite{yi2020applying}  \revision{or domain-specific tasks~\cite{zuluaga2023does}}. \revision{Pre-trained wav2vec2.0 models have been made available by researchers on the Huggingface platform.\footnote{\url{https://huggingface.co/models?search=wav2vec2.0}}}

Wav2vec2.0 has shown improvements compared to the prior state of the art in ASR, which \revision{were} Recurrent Neural Networks. At the time of writing, the state of the art on common ASR benchmarks is held by Wav2vec2.0 models.\footnote{\url{https://paperswithcode.com/task/speech-recognition}, accessed in February 2023} \revision{Yi et al. \cite{yi2020applying} showed that Wav2vec2.0 is particularly successful for low resource languages, where training data is scarce. They report more than 20\% relative improvement for six languages compared with previous methods.}

%A model that is finetuned towards ASR can be compared to an acoustic model in classic ASR in the sense that it turns a speech signal into character probabilities. These are then decoded to generate a text. 
%By default the Huggingface implementation uses a greedy decoder, which outputs the characters with the highest probability.\footnote{At the time of writing, language model decoding has been integrated in Huggingface. This was not the case when we implemented our solution. \url{https://huggingface.co/docs/transformers/v4.14.1/model_doc/wav2vec2\#transformers.Wav2Vec2ProcessorWithLM}} 

%\subsection{Prior work with Wav2vec2.0}

%As the basis for our methodologies we use Wav2vec2.0 \cite{baevski2020wav2vec}, a Transformer-based ASR method that uses a self-supervised pre-trained model to learn representations of sound files. The main hurdle that needed to be taken for this to be possible was to quantize the analogous signal to discrete units for which the authors proposed vq-wav2vec (vector quantization wav to vec) \cite{baevski2019vq}. 

%Similar to Transformer models for text data, 

Wang et al \cite{wang-etal-2021-voxpopuli} have fine-tuned a Wav2vec2.0 model %speech representations can be fine-tuned to domain-specific ASR tasks using manually transcribed data. A particularly relevant fine-tuned model is the Voxpopuli model, which was finetuned 
on speech and text from the plenary sessions of the European Parliament. We use the Voxpopuli model as \revision{a} domain-specific acoustic model in our experiments.

\subsection{Related work in Political Science}

\revision{Corpus compilation in the political domain is an active area of work. The most prominent initiative in this context is the CLARIN project ParlaMint.\footnote{\url{https://www.clarin.eu/parlamint}} The aim of ParlaMint is the the development and release of comparable and interoperable corpora of the national parliaments in Europe. The corpora consist of transcriptions of parliamentary speech. The project has been successful, publishing 17 corpora until 2021~\cite{erjavec2022parlamint} that have been uniformly encoded and enriched with metadata and automatically generated linguistic annotations. The corpus development in the ParlaMint project starts with transcriptions of speech, not with the raw speech input. That means that ASR is outside the scope of their work.}

\revision{In other work i}n the political domain, the use of ASR is often mentioned as \revision{a promising} method for creating a corpus, yet the amount of studies actually pursuing this is rather limited. 
The only recent study that investigates the use of ASR to generate textual data in the political domain is the work by Proksch et al. \cite{proksch_testing_2019}. 
This article has been cited in a number of studies (such as \cite{munger2022right}, \cite{wratil2019public} and \cite{schoonvelde2019text}) as having potential (often as a direction of future research), but none of the papers that cite it have used the method.\footnote{According to Google Scholar as of April 2022}

% Corpus building with ASR

% Hard to find ASR in the political domain.
% \begin{itemize}
%     \item Proksch et al. Also discussed below.\cite{proksch_testing_2019}
%     \item Defrancq et al. \cite{defrancq2021automatic} study the use of ASR to aid the live interpreting. Also use the google speech to text API as Proksch et al. do as well.
% \end{itemize}

%Thus, there is little prior work employing ASR in the fields of Political Science and Public administration. The most prominent study is the work by \suzan{Proksch et al.} \cite{proksch_testing_2019}. 
Proksch et al. \cite{proksch_testing_2019} evaluate two ASR services (Google API and Youtube subtitles) for their usefulness in political research. They perform an intrinsic evaluation by calculating the Word Error Rate (WER), a metric that measures the rate of incorrectly transcribed words by comparing the automatic transcription with a manual transcription. They also perform extrinsic evaluation, in which they analyze the suitability of the generated transcriptions as input for Wordfish \cite{slapin_scaling_2008}, \revision{a text processing tool\footnote{\url{http://www.wordfish.org/}}} that is commonly used in political science to infer the position of actors on a right--left scale based on texts they have produced. 

As intrinsic evaluation, Proksch et al. report WERs between 0.03 (YouTube subtitles)\footnote{This API is not available to experiment with.} and 0.26 (Google API) for English \cite{proksch_testing_2019}. Based on the extrinsic evaluation, \revision{moreover,} the authors conclude that there is hardly \revision{any} difference between using \suzan{human-transcribed} text and automatically transcribed text as data source for Wordfish; Both yielded similar results. From this they conclude that the Google API and the YouTube API are useful for generating transcriptions that are used in bag-of-words approaches to text analysis. 
%In our case we used a model that could be be optimized for a specific domain. 
We showed in previous work~\cite{deVos2020challenges} that general \revision{open-source} ASR models \revision{-- in our case the English Automatic Speech Recognition Webservice based on the Kaldi
framework --} are limited by the mis-recognition of names and domain-specific terms in the political domain. Therefore, we focus on domain adaptation of ASR in this paper. Our main contribution compared to Proksch et al. is that we address domain adaptation of ASR for the political domain. We investigate the effectiveness of using a domain specific acoustic model, using a domain specific decoder language model, and the addition of domain-specific \revision{terms to the ASR decoder}.  Our solution uses open-source code, which means that our pipeline can be run locally and does not depend on a third party API such as Google's speech to text. %In the scientific context this has the advantage of reprehensibility as well as that it can be run on own servers in case privacy or other sensitivities with regard to the data are an issue. 

\section{Data}\label{sec:corpusdescription}

\subsection{Audio}\label{subsec:corpusaudio}

The sound files of the LIBE committee \revision{were} downloaded from the webstreaming service\footnote{\url{https://www.europarl.europa.eu/committees/en/meetings/webstreaming}} of the EU parliament. In total we downloaded 332 files, with a mean length of 152.63
% of in seconden weergeven? Of minutes:seconds instead of tens and hundreds.
minutes (sd: 60.50). The shortest file was 9.18 minutes and the longest was 289.43 minutes.
These audio files constituted the entire parliamentary session from 2014 to 2019.
% Deze mededeling is mss niet relevant? Maar de URL moet wel ergens genoemd worden denk ik.
Originally the files are in mp4 format.\footnote{The files are stored as mp4 with a black screen as video content.} We converted them to 16kHz 16 bit PCM wav with ffmpeg as input for the ASR. %\footnote{The command we used was: ffmpeg -i soundfile.mp4 -f wav -vn soundfile.wav. For more info, see: \url{https://www.ffmpeg.org/}}
%, as the ASR software requires this as input. 
% ffmpeg -i <soundfile>.mp4 -f wav -vn <soundfile>.wav

% In total we transcribed 202 audio files with a total length of 547 hours and 24 minutes (547.396 hours). 
% The shortest audio file was 9 minutes and the longest 4 hours and 31 minutes.

% We transcribed a total of xx words. On average xx words per meeting (sd: xx).

\subsection{Manually transcribed evaluation set}\label{subsec:manualset}
\revision{For evaluation purposes, we created a small set of manually transcribed data. Note that we don't use this set for training or finetuning our ASR models -- which were pre-trained on different data -- but only for evaluating different models for our data and optimizing the hyperparameters in a cross-validation setting. The evaluation set consists of} 100 randomly selected samples with an average length of 21.57 seconds (sd 4.41). The samples \revision{were} randomly selected according to the following procedure. 
First, we randomly selected 10 meetings. Then from each meeting we randomly selected a segment $s$ (with the only constraint that there \revision{were} at least 9 segments after $s$). Then this segment and the 9 after it \revision{were} selected to be part of the evaluation set. This way, from ten random meetings, we randomly selected ten consecutive segments that we then manually transcribed. This lead to \revision{a} total of 4,893 words of transcribed text which constituted 36 minutes and 18 seconds of audio material.

\subsection{Metadata}\label{subsec:metadata}

For the purpose of collecting domain- and context-specific terms (see Section \ref{subsec:experiment3}), we downloaded agendas and minutes of each meeting. In total we downloaded 432 metadata files: 235 agendas and 197 minutes. They where downloaded by hand from the official website of the EU-parliament. 
On average, the agendas contained 483.27 (sd: 365.04) words and the minutes contained 2088.93 (sd: 1123.94) words.

It must be noted that the audio is not always aligned with the metadata. The reason for this is that the audio can be retrieved per session and the agenda and minutes are uploaded for a meeting, which consists of multiple sessions and can span multiple days.

We therefore did a semi-automatic alignment between the audio and the metadata.
For every audio file, we first checked whether there was an agenda and minutes that were dated at the same day as the audio file. If this was the case, these were automatically linked. If not, we manually evaluated all documents within a window of 5 days before and 2 days after the date of the audio file, and if any of these documents corresponded with the meeting, they where connected.
It could also occur that we could not retrieve any document for a session. Then we concluded that there was no metadata for this session in the online archives of the EU parliament and we stored the audio file without metadata.

\section{Methods}\label{sec:methods}

\subsection{Experimental Design}

\begin{figure}[h]
    \centering
    \includegraphics[width=0.7\textwidth]{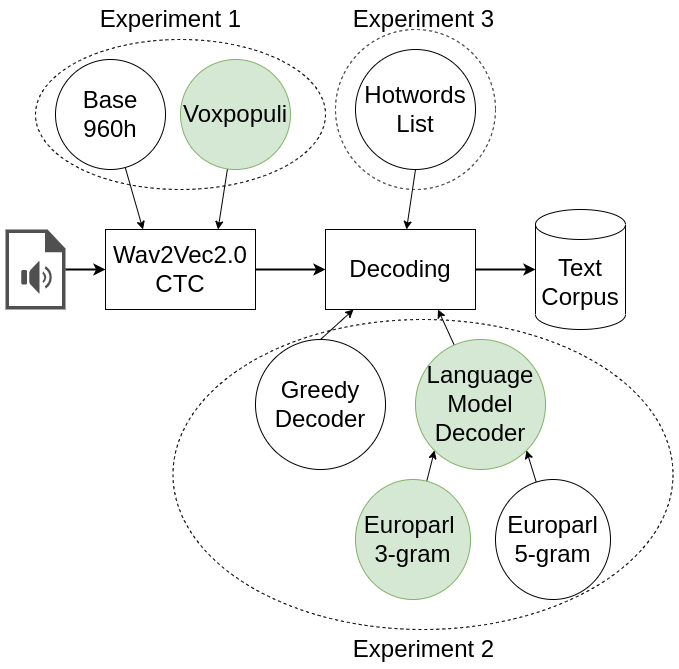}
    \caption{Schematic overview of the experimental design. The green elements are the elements that gave the best result and were used in creating the corpus.}
    \label{fig:mesh1}
\end{figure}

Figure~\ref{fig:mesh1} illustrates our experimental design. It shows that we experiment with different settings in multiple stages of the ASR pipeline: Experiment 1 where we intervene in the Wav2Vec2.0 acoustic model and Experiments 2 \revision{and 3} where we intervene in the decoding stage. The green modules are the ones that lead to the best performance and therefore were used in creating the corpus.

As acoustic models we compare wav2vec2--base--960h, wav2vec2--large--960h, and wav2vec2--base--10k--voxpopuli--ft--en. 
For the decoding, we compared a greedy decoder, beam search decoder without language model, and beam search decoder with language model. 
In addition, we experiment with the use of domain- and context-specific terms as so-called hotwords: a list of words added to the decoder module that receive a boost in the recognition.

\subsection{splitting the audio}\label{subsec:splitting}

For \suzan{processing by} Wav2vec2 the recommended length of an audio file is 10--30 seconds.\footnote{\url{https://github.com/pytorch/fairseq/tree/main/examples/wav2vec##training-a-new-model-with-the-cli-tools}}
For this reason we split the files in segments of such length. 
Existing functions for splitting audiofiles did not work sufficiently on our data. Most functions (for example the librosa function\footnote{\url{https://librosa.org/doc/main/generated/librosa.effects.split.html}}) work by optimizing a `loudness' threshold below which everything is considered to be silence and therefore a breaking point. 
The problem with our data was that there is a huge variety within an audio file between speakers and the noisiness of their microphone. As a result, some speakers never got below the silence threshold and others often. This resulted in a lot of audio segments of 1 second and also some of 20 minutes (but averaging at 20 seconds).

\suzan{As a solution, we developed} an algorithm that \suzan{seeks} local solutions. It \suzan{iteratively takes} the most silent moment (lowest average absolute amplitude) between 5 and 30 seconds after a starting point and \suzan{defines} that moment as the break point. After that the breaking point is the new start point. 
This was done by moving a window of 1 second with a step size of 0.1 second between 5 and 30 seconds after the starting point. Every window consisted of 16k amplitude values (because of the sample rate of 16k). For every window, the sum of the squared amplitude values was calculated. The center of the window with the lowest sum of squared amplitudes was determined to be the cutoff point.
% asdf

\suzan{The data splitting led to 128,918 segments with an average length of 21.64 seconds (sd: 5.72).}

\subsection{Baseline setting}\label{subsec:experimentBaseline}

As baseline we use a zero-shot run of Wav2Vec2.0 as available in Huggingface with the default parameter settings, the Wav2Vec2-Base-960h model and the standard greedy decoder. 
Besides the baseline experiment, we also tried the Google cloud speech-to-text API. 
The main motivation behind this was to compare with an often used product as well as to get a comparison to the paper by Proksch et al. \cite{proksch_testing_2019}. The results of these experiments are included in Table \ref{tab:wer_results} as Baseline 1 \& 2.

\subsection{Experiment 1: wav2vec2-base vs voxpopuli}\label{subsec:experiment1}
In this experiment we compare 2 pretrained models for wav2vec2.0:
Wav2Vec2-Base-960h
\footnote{\url{https://huggingface.co/facebook/wav2vec2-base-960h\#wav2vec2-base-960h}} 
and wav2vec2-base-10k-voxpopuli-ft-en
\footnote{\url{https://huggingface.co/facebook/wav2vec2-base-10k-voxpopuli-ft-en}}.
The former was trained and finetuned on the LibriSpeech data set \cite{panayotov2015librispeech}, which comprises 960 hours of audiobooks and the latter (Voxpopuli) was pretrained on 10K hours from an unlabeled subset of the VoxPopuli corpus and fine-tuned on 543 hours of transcribed speech from the same corpus \cite{wang-etal-2021-voxpopuli}. 
The voxpopuli corpus is a corpus of speech from plenary sessions of the EU parliament. 

%The idea/hypothesis is that the Voxpopuli model would work better, because the data it was trained on was more similar to the data that we applied it on in terms of topic domain, speaker variety, and microphone quality.

% For both models we used the default wav2vec2.0 hyperparameters: %We ran the default wav2vec2.0 pipeline with both models on the evaluation set, keeping all settings except for the used model the same. 

\subsection{Experiment 2: Decoding with a language model}\label{subsec:experiment2}

By default the Wav2Vec2.0 implementation of Huggingface does not use a language model to decode the logit vectors that are the output of the Wav2Vec2.0 CTC. 
We used the pyctcdecode package\footnote{\url{https://github.com/kensho-technologies/pyctcdecode}} to be able to use knowledge from a language model to decode the logits. 
For fitting the language model we used the kenlm \cite{heafield_kenlm_nodate} software\footnote{See also \url{https://kheafield.com/code/kenlm/}} and we fitted the models on the English part of the Europarl corpus \cite{koehn_europarl_nodate}. Fitting in this context means estimating the conditional probabilities of each word as well as the backoff probabilities. %with the kenlm software \cite{heafield_kenlm_nodate} .

We experiment with 3 configurations: pyctcdecode without language model, 
pyctcdecode with a 3-gram language model, and pyctcdecode with a 5-gram language model.
For ctc decoding, the alpha and beta hyperparameters need to be tuned. We did this according to the cross validation procedure \revision{on the evaluation set} described in section \ref{subsubsec:crossvalidation}.

% kenlm commands: 
% kenlm/build/bin/lmplz --text europarl_for_kenlm.txt --arpa europarl-5-gram.arpa -o 3 -T temp/
% bin/build_binary europarl-3-gram.arpa europarl-3-gram.bin

Before fitting the language model, we applied some preprocessing steps: 

\begin{itemize}
    \item Removing all xml tags. (anything between \textless~and \textgreater)
    \item Removing all text within parentheses.
    \item Sentence splitting such that there is 1 sentence per line. For this we used the NLTK sentence tokenizer\footnote{\url{https://www.nltk.org/api/nltk.tokenize.html}}.
    \item Replacing all numbers with words. (e.g. `1' $\rightarrow$ `one') For this we used the num2words package in python\footnote{\url{https://pypi.org/project/num2words/}}.
    \item Lowercasing
    \item Removing all non-word characters except for the apostrophe. This was done by replacing them by whitespaces after which any double spacing was removed.
    \item Tokenizing the apostrophe. i.e. regarding the apostrophe as a separate token by placing it between whitespaces. This was done to make the model more consistent with the wav2vec2.0 tokenization.
\end{itemize}

% (refer to Table 1 of paper for more information).

\subsection{Experiment 3: Hotwords}\label{subsec:experiment3}

The main challenge in domain-specific ASR is the recognition of domain terms and named entities \cite{deVos2020challenges}. To address that challenge, Pyctcdecode provides the possibility of hotwords boosting. With this function a list of hotwords can be added. These words get boosted and therefore \textcolor{black}{are more likely to be recognized}. Together with the hotwords, a weight must be provided that determines the priority given to \revision{them}. With a higher weight, the decoder will recognize more hotwords at the cost of other words.

In the political domain, names of persons and organizations that play a role in the decision making process are potentially relevant terms. Recognizing those correctly will make the corpus more \revision{useful} for political researchers. We therefore compiled hotwords lists as follows. We downloaded all the agendas and minutes of the meetings. \revision{As reported in Section~\ref{subsec:metadata}}, we had agendas for \textcolor{black}{235} of the meetings and minutes for \textcolor{black}{197} of the meetings. Sometimes documents could not be retrieved as they where unavailable in the repository of the EU parliament.
We automatically extracted entities from the meeting documents with \revision{the named entity recognizer provided by the Spacy NLP library}.\footnote{\url{https://spacy.io/api/entityrecognizer}} 
Of all the entity types Spacy can recognize, we included organizations (ORG), persons (PERSON), geo-political entities (GPE) and locations (LOC), because these are the entities that are most valuable in the political context. 
We included the meeting-specific hotwords during the transcription of each meeting. In the test set (see Section \ref{subsec:manualset}), there was one meeting for which no agenda and minutes were available in the EU repository.

\paragraph{Hotwords evaluation}\label{subsubsec:PrecisionRecall}

%Besides the WER, we also looked at the precision and recall at which entities were recognized. This procedure is described in \ref{subsubsec:PrecisionRecall}.

To evaluate the general quality of the ASR we report the Word Error Rate (WER) (see Section \ref{subsec:methods-evaluation}). Besides that, we used precision, recall and F-measure to evaluate the recognition of the hotwords. To calculate the measures we used specific definitions of True Positives, False Positives and False Negatives on the token level.
\begin{itemize}
\item True positives (TP) are entities that are correctly transcribed by the ASR. This includes the terms that are in the hotwords list, the reference text as well as the transcribed text. 
\item False positives (FP) are entities that are in the hotwords list and the automatically transcribed text but not in the reference text. 
\item False negatives (FN) are the terms that are in the hotwords list and the reference text, but not in the automatically transcribed text.
\end{itemize}
Contrary to the WER, which was calculated for every segment, the precision recall and F-score where calculated per meeting: We counted all TPs, FPs and FNs of all segments in a meeting and computed the evaluation metrics with those.

\subsection{Evaluation}\label{subsec:methods-evaluation}

\paragraph{Evaluation metric}
We use the commonly used Word Error Rate (WER) to evaluate our models. The WER is defined in equation \ref{eq:WER} where S is the number of substitutions, D the number of deletions, I the number of insertions and N the number of words in the reference text.
% Dit komt vrij letterlijk van wikipedia. Even kijken of ik dit anders moet verwoorden. https://en.wikipedia.org/wiki/Word_error_rate 

\begin{equation}\label{eq:WER}
    WER = \frac{S+D+I}{N}
\end{equation}
\\
We use the jiwer\footnote{https://github.com/jitsi/jiwer} package to calculate the WER for every sentence in the annotated set and then compute the average WER over all sentences.

% In experiments where hyperparameters need to be tuned, we used 5-fold-cross validation to tune the hyperparameters. 
% We created the the folds in such a way that the evaluation fold always consisted of 2 of the 10 transcribed meetings such that different segments from the same meeting did not end up in different folds.

% \begin{figure}[h]
%     \centering
%     \includegraphics[width=0.25\textwidth]{figures/cross_validation.png}
%     \caption{The cross validation setup.}
%     \label{fig:crossvalidation}
% \end{figure}

\paragraph{Cross Validation}\label{subsubsec:crossvalidation}
When hyperparameters needed to be tuned (experiment 2), we used 5-fold cross validation \revision{on the manually transcribed evaluation set (see Section~\ref{subsec:manualset}}). We \revision{split the data in such a way} that sound files from the same meeting \revision{are in the same partition, so that there is no information leak caused by overlapping topics or speakers between the train partitions and the test partition}. \revision{With 100 segments in our manually transcribed set, each partition contained 20 segments. }
%This means that the evaluation fold contained 20 segments from 2 meetings. And that from those meetings, no segments were part of the tuning set.
%This was done to get the best estimate of how the model would perform on unseen data.

\paragraph{Preprocessing for evaluation}
For the evaluation, \suzan{the same} preprocessing steps were applied to both the ASR output as well as the manual transcriptions. The preprocessing steps applied were:

\begin{itemize}
    \item Lowercasing
    \item Removing punctuation except for the apostrophe.
    \item Replace diacritics to their non-diacritic equivalent.\footnote{The reason for this is that the English Wav2Vec2.0 model did not have diacritics in its vocabulary.} In our data, this only applied to the word \textit{Orbán} (the name of the Hungarian president), which was changed to orban.
    \item Remove double spacing 
    \item Remove \revision{hesitation} words/sounds. These words where inconsistently \suzan{transcribed} by the ASR but they were consistently transcribed \suzan{manually} as `eh'. This unnecessarily inflated the WER. Words we removed are: `eh', `ehm', `er', `ar', `uh', `e', `eh'
\end{itemize}

% Precision was defined as:

% \begin{equation}\label{eq:Precision}
%     P = \frac{TP}{TP + FP}
% \end{equation}
% \\

% Recall was calculated as:

% \begin{equation}\label{eq:Recall}
%     R = \frac{TP}{TP + FN}
% \end{equation}
% \\

% The F1 was calculated as:
% \begin{equation}\label{eq:F1}
%     F1 = 2 \times \frac{P \times R}{P + R}
% \end{equation}

\section{Results}\label{sec:results}

\subsection{ASR results}

All experimental results are shown in Table \ref{tab:wer_results}.
%\paragraph{Google Cloud API}
The table shows that the \textbf{Google Cloud API }(Baseline 1) performs poorly on this particular data. \revision{Analyzing this further, w}e noticed that the quality of the output of the Google API strongly \revision{varies between} utterances. In some occasions it \revision{performs} well (see Example \ref{ex:googleAPIgood}) but in other samples it seems to have recognized only a few words (see Example \ref{ex:googleAPIbad}). 

%% Example with parts
\pex<googleAPIgood>\label{ex:googleAPIgood}   %% Longest judgement mark
% first part
\a {[ref]}\footnote{``ref'' indicates the human transcribed reference text and ``asr'' the automatically transcribed text text.} I would like to stress that the cooperation between our committees is very important because there are areas of common interest among us, there are also areas of responsibility of the commissional designate
% second part (with judgement}
\a {[asr-base1]} I would like to stress that in the cooperation between our Communities is very important because there are areas of common interest among their also areas of responsibility of the commissioner designate 
\xe
% \begin{exe}
% 	\ex This is a numbered example.
% \end{exe}

\pex<googleAPIbad>\label{ex:googleAPIbad}   %% Longest judgement mark
% first part
\a {[ref]} remarks, and I would also like to know if the commission continues to refuse to propose an lgbti roadmap. And I would really like to know what the commission will do, finally, to start protecting the privacy of European citizens, and I conclude
% second part (with judgement}
\a {[asr-base1]} like to know 
\xe

This result is different from the result presented in Proksch et al. \cite{proksch_testing_2019} who report a WER of 0.26 for the google API. Looking at the data appended to their paper, the omission of large parts of texts seems not to be an issue in their case. In our data, the quality of the transcriptions seems indeed acceptable, in the parts where it did not omit sentences. \revision{In that sense, the WER score for the Google Cloud API is not very meaningful for our data, because of the parts omitted by the recognizer. We speculate that the omissions might have been caused by variable recording qualities, the use of non-native English, or the combination of speech by the original speaker and the interpreter. }

The \textbf{wav2vec base} setting already shows a substantial improvement compared to the Google Cloud API, and also compared to our earlier pilot study \cite{deVos2020challenges} using the KALDI framework, although that latter was evaluated on a different test set. 
In general, the results from the Baseline 2 setting are reasonable, but it also produced non-words such as `resa' in example \ref{ex:ex0}. This example also exhibits an effect that is the result of the way interpreters are present in the audio: If a speaker does not speak English, the first few seconds of their speech is uttered in the original language before the English interpreter replaces the speaker's voice. This effect results in a few seconds of gibberish in the output of the ASR.

\pex<ex0> \label{ex:ex0}
\a {[ref]}
yes i would like to raise a point regarding procedure
\a {[asr-base2]} operch culat that's it innkemured yes i would like to resa point regarding procedure
\xe

\begin{table}[t]
\begin{tabular}{llr}
\textbf{Experiment}               & \textbf{Configuration}                & \textbf{WER} \\ \hline
\multicolumn{1}{l|}{Baseline 1} & Google Cloud API & 73.46    \\ 
\multicolumn{1}{l|}{Baseline 2} & wav2vec2.0 base                                    & 28.22        \\
\multicolumn{1}{l|}{Experiment 1} & wav2vec2.0 voxpopuli                             & 21.42        \\
\multicolumn{1}{l|}{Experiment 2} & wav2vec2.0 voxpopuli + language model            & 17.95        \\
\multicolumn{1}{l|}{Experiment 3} & wav2vec2.0 voxpopuli + language model + hotwords & 25.53 
   \\      

\hline
\end{tabular}
\caption{\label{tab:wer_results} The best result of each experiment (lower is better). \revision{The WER for Baseline 1 is artificially low because parts of utterances were omitted in the transcript of the Google Cloud API. Baseline 2 is the wav2vec2.0 model used in a zero-shot setting. }}
\end{table}

Experiment 1 shows that the use of the domain-specific \textbf{Voxpopuli model} leads to a reduction of the WER of 6.80 \revision{percent} points compared to the base model (\revision{baseline 2}). For example, ``raise a'' from Example \ref{ex:ex0} was recognized correctly (see example \ref{ex:ex1}\revision{)}. Remarkably, the output of experiment 1 also did not contain the gibberish. We cannot say exactly say why this is the case, but we suspect that during finetuning on data of the plenary sessions of the European Parliament, where interpreters are used in a similar fashion, the model has learned to ignore everything that is not English. This ignoring of non English is a big source \revision{of} improvement of the WER. Interestingly, `yes' is not recognized in Example \ref{ex:ex1}. A likely cause for this is that the transition between the original language and English is very sudden and that yes is not recognized as part of the English utterance. 

\ex<ex1> \label{ex:ex1}
{[asr-voxp]} s i would like to raise a point regarding procedure
\xe

The replacement of the greedy decoding model of the wav2vec2.0 implementation by decoding with a \textbf{language model} in experiment 2 leads to a reduction of the WER to 17.95 after the alpha and beta parameter were tuned. This is an improvement of 10.27 percent points compared to the baseline and 3.47 compared to experiment 1. 
As a result, the abovementioned example was recognized perfectly, %with the exception of capital letters and punctuation 
resulting in Example \ref{ex:perfectscore}

\ex<ex3>\label{ex:perfectscore}
{[asr-lm]} yes i would like to raise a point regarding procedure
\xe

\subsection{Hotwords Experiment results}\label{sec:hotwordsexperimentresults}
The weight parameter \revision{for the hotwords} was optimized using ten fold cross validation using the same folds as described in Section \ref{subsubsec:crossvalidation}. We optimized the weight parameter \revision{for recall} on the tuning partition and then applied it to the test partition. \revision{We opted for recall as optimization metric in order to investigate the model quality for a case in which named entities are important to recognize, e.g. searching the data for named entities or analyzing stance towards political entities (a common goal in the political domain).} %The results of the tuning are \revision{discussed in Section \ref{sec:hotwordsexperimentresults}}.

\begin{figure}[t]
    \centering
    \includegraphics[width=0.8\textwidth]{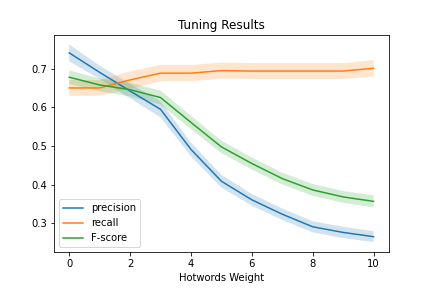}
    \caption{The mean results of the hotwords experiment as described in section \ref{subsec:experiment3}. The shaded area depicts 1 standard deviation from the mean.}
    \label{fig:hotwordsresults}
\end{figure}

Figure \ref{fig:hotwordsresults} displays the results of tuning the weight for the hotwords list on the validation set during decoding. This experiment is explained in \textcolor{black}{Section \ref{subsec:experiment3}}. 
This figure shows that the optimal \revision{recall for hotwords on the development set is obtained with the weight value} 3. With this value, the recall for hotwords \revision{in the test set} is 0.85, compared to 0.82 with no hotwords weight. \revision{With a hotwords weight smaller than 3, recall of the hotwords recognition will be lower, and the ASR will recognize fewer entity names. The hotword weight is a trade-off between recognizing named entities and the overall WER: Table~\ref{tab:wer_results} shows that the overall WER is lower with hotwords than without hotwords. This is because hotwords are relatively infrequent words and when including hotwords in the model (Experiment 3), more frequent words (function words like `you') are sometimes incorrectly recognized as hotword (`eu'). Without hotwords (Experiment 2), more words are transcribed correctly (leading to a lower WER), but entities are more often missed by the ASR. } With a hotwords weight larger than 3, the false positives start to outweigh the gain in true positives: when the weight is larger, the decoder is more prone to recogniz\revision{ing} regular words as hotwords\revision{, and} the false positives \revision{more heavily affects} the overall WER. 

An example of a false positive is presented in Example \ref{ex:falsePositive}. In this example, `you' is recognized as `eu'. This is an example where the domain-specific hotword EU was over-recognized and \revision{used to replace} non-entity words. In manual exploration of the results, we saw that most often, a function word was mis-recognized to be a hotword. 

\pex<falsepositive>\label{ex:falsePositive}
\a {[ref]} I am glad somebody raised the point, I think the substantive point \textbf{you} are making is
\a {[asr-hotwords]} (...) iam glad to somebody raised the point i think the substantive point \textbf{eu} are making (...)
\xe

Table~\ref{tab:hotwords_further_analysis} shows the result of the hotwords experiments on the test set in terms of precision and recall for the hotwords, and WER for the complete transcription. We show the results for the 10 individual meetings. This table shows the difference in quality per meeting. The difference between the highest and the lowest WERs is 7.49 percent points. There are also large differences among precision, recall, and F-measures for different meetings. This pattern is illustrative for the other experiments as well, where there were also large individual differences between the individual meetings. \revision{There is no significant correlation between WER and F1 (Spearman's $\rho=0.143, p=0.71$), indicating that the quality of the hotwords recognition does not correlate to the quality of the speech recognizer in terms of WER. }

\begin{table}[t]
\begin{tabular}{l|rrrr}
\toprule
meeting identifier & precision & recall &  F1 &   WER \\
% &           &        &     &       \\
\midrule
20140714           &       .90 &    .66 & .84 & 24.99 \\
20140723           &       .95 &    .95 & .93 & 21.06 \\
20140904           &       .81 &    .92 & .85 & 25.48 \\
20140915           &       .76 &    1.00 & .84 & 26.10 \\
20140925           &       .58 &    .74 & .78 & 26.17 \\
20141016           &       .82 &    .95 & .86 & 24.11 \\
20141106           &       .81 &    .90 & .93 & 26.26 \\
20141110           &       .74 &    .80 & .88 & 28.55 \\
20141120           &       .51 &    .60 & .70 & 21.22 \\
\bottomrule
\end{tabular}
\caption{\label{tab:hotwords_further_analysis} Comparison of the average results per meeting. This shows that there is a large variation between meetings. One meeting was omitted in this table as there were no hotwords for this meeting \revision{and} as a result of that there were no documents for this meeting.}

\end{table}

% \subsection{Error analysis}\label{subsec:error_analysis}
% In table \ref{tab:outputexamples} we present a few examples of the output. These examples show that the quality of the output varies between segments. 
% The first example (20140723\_068) is an example in which an interpreter is speaking. In general the best transcriptions are achieved for the interpreters. This is likely because interpreters speak slowly and clearly and often have no big accent. They also tend to create better sentences.
% The second example 20140915\_002 is another example of a good transcription. This is an example of a member of parliament who spoke English with only a small accent.
% The third example (20140701\_020) is an example of cases where the transcription was very bad. 
% In these cases, the decoder could not identify spaces. This issue was largely solved by optimizing the alpha and beta parameter. 
% Yet even the optimal values still left a few cases like these. In the transcribed dataset there are 2 examples (including this example) of texts that consist entirely of such "clumped together" strings.

% The fourth (20140701\_016) and fifth (20140925\_041) examples are more typical results, that are not perfect but are readable. Despite not using a hotwords list, some of the named entities such as Portugal and Hungary are recognized correctly yet others such as Martin Schultz (martin jules) are not. 

% These examples show that, despite the texts are not perfect, they often are readable and useful.

\subsection{The resulting corpus}
Based on the results of experiments 1-3 we created a corpus with the voxpopuli model and a decoder that used a 3-gram language model that was fitted on the English part of Europarl: the settings from experiment 2.

\revision{The addition of hotwords} lead to a substantially better recognition of named entities \revision{(see Section~\ref{sec:hotwordsexperimentresults})}, something which is useful in the political context, it also lead to a large decrease in overall WER \revision{(Experiment 3 in Table~\ref{tab:wer_results})}. %Because the precision and recall are far from perfect, this also means that the transcriptions cannot be reliably used for analyses on the number of occurrences of Named Entities. 
Everything considered, we decided that the advantages of using hotwords, did not weigh up against the decrease in transcription quality.

% \subsubsection{General overview and statistics}
The resulting corpus consists of $3,584,150$ words. On average the 10-30 second segments contained $45.02$ words (sd: $8.78$). 
The mean number of words at a meeting was $18,401$ (sd: $7,681$). 
Given an average meeting length of 152 minutes, this means an average of 7263.55 words per hour. %which is less than for example the Corpus Gesproken Nederlands (Corpus Spoken Dutch) \cite{oostdijk2000het} which contains more natural conversations and has an average of about 10,000 words per hour.

\subsection{Corpus exploration through topic modeling}\label{subsec:topicmodeling}

To get a general overview of the contents of the corpus, we fitted an LDA-topic model \revision{\cite{blei2003latent}} with 8 topics. The number of topics was determined by selecting the number of topics for which the average topic coherence was the highest according to the word2vec based coherence metric by O'Callaghan and Greene \cite{oCalaghan2015analysis}. In this method, the average cosine distance between the vectors for the top 15 words within a topic is minimized. 

Before topic modeling we applied a number of preprocessing steps. For stop word removal we used the English stopword list from NLTK\footnote{\url{https://www.nltk.org/book/ch02.html}}. We lemmatized the corpus with the NLTK WordNetLemmatizer.\footnote{\url{https://www.nltk.org/_modules/nltk/stem/wordnet.html}} %We chose lemmatization over stemming to make the topics easier to interpret. 
Lowercasing was not necessary because the output of the ASR is already lowercased. 

The LDA output gave an insightful mix of substantial as well as procedural topics. An overview of all topics can be found in appendix \ref{appsec:topics}.

\paragraph{Procedural topics}\label{para:procedural_topics}
Examples of procedural topics are topics about voting and starting/ending a speech. We identify two clearly procedural topics:
\begin{itemize} \item Topic \#4: the top 10 words of this topic contains words such as \textit{vote, open, amendment, closed, rejected, adopted}. This is indicative of the voting procedure in which the chair first announces that the voting is opened (the MEPs can cast their votes via the electronic voting system), then that the votes are closed followed by the announcement of the result: ``the amendment is adopted/rejected.'' \item Topic \#7: this topic has high-probability words about starting and ending a speech such as \textit{mr, thank, president, much, colleague, rapporteur}. This is indicative of the way speakers start and end their speeches: by thanking everyone (the president, colleagues, rapporteurs) for their presence and contributions. \end{itemize}

\paragraph{Substantive topics}\label{para:substantial_topics}
Clear examples of substantial topics are topics about data protection and refugees. 
\begin{itemize}

\item Topic \#1 (\textit{crime, terrorism, organised, victim, fight, human, trafficking}) is a topic on the fighting of organized crime such as human trafficking and terrorism. The prevalence of this topic can be explained by the fact that in 2016 the LIBE committee took over the work of the CRIM committee (which was fully dedicated to organized crime) which was cancelled in that year. \footnote{\url{https://www.europarl.europa.eu/committees/en/fight-against-organised-crime-and-corrup/product-details/20160216CHE00191 }}

\item Topic \#3 (\textit{right, data, protection, information, privacy}) can be explained by the fact the the General Data Protection Regulation (GDPR) was discussed in parliament during this parliament and that the LIBE committee was the committee that was mainly responsible for this legislation.

\item Topic \#6 (\textit{border, country, refugee, migration, turkey, agreement}) can be explained by both the fact that refugees rights are one of the main topics of interest of the LIBE committee as well as the fact that the European refugee crisis, as well as the agreement with Turkey which was an attempt to stop it, took place in 2015, a year that is included in our sample.

\item Topic \#8 (\textit{two, thousand, year, hundred, one, twenty}) mainly consists of numbers. Based on the two most prevalent words being \textit{two} and \textit{thousand} it is likely that numbers are mostly used to express dates and times.

\end{itemize}

\section{Conclusion}\label{sec:Conclusion}
% (1) We show that using a domain-specific acoustic model and an adapted language model improves the quality of transformer-based speech recognition in the political domain as well as how much gain the different adaptations can provide;
% (2) we deliver a corpus of transcribed LIBE meetings as well as a pipeline that can be used to transcribe EU parliament meetings.\footnote{All our code can be found at

In this paper we experimented with different pipelines of the wav2vec2.0 ASR models to automatically create a text corpus from the meetings of the LIBE committee of the European parliament. Topic modeling indicates that the corpus is of sufficient quality to perform downstream analyses. \revision{We also showed that the ASR of the Google Cloud API is not sufficiently robust to transcribe our noisy data: Although the transcribed sections are of good quality -- in line with prior work~\cite{proksch_testing_2019} -- large parts of utterances are sometimes omitted in the ASR output, likely because of the low quality of the recording combined with the non-English speech in the background of the interpreter's English. For high-quality recordings and direct, zero-shot use, the Google Cloud API is probably a good choice, but for more noisy and domain-specific data, wav2vec2.0 is a better option because it is adaptable to the domain language and more robust against noisy speech and lower-quality audio.}

We showed that the quality of the corpus can be improved by using a domain specific acoustic model as well as using a language model to decode the output. %However, we must note that we were lucky that a large pretrained model already exists for the European Parliament, namely the Voxpopuli model. So using a domain specific acoustic model for other domains might be easier said than done.
%With this, we showed that it is feasible to create a political text corpus using existing tools without going through the expensive task of creating a corpus that is large enough for finetuning a wav2vec2 model. 
%One important condition for this is to have a large pretrained model available for the desired domain. Something that was there for the European Parliament, but this might be an exception.

The recognition of domain-specific terms can be improved by hotwords boosting. In our case we were able to recognize more domain-specific entities (names), but at the cost of the overall performance in terms of word error rate. We decided that we would get the best generic corpus \revision{in terms of WER} without using hotwords, but\revision{, as argued by Favre et al. \cite{favre2013automatic},} for a corpus that is made towards a specific research goal\revision{, the priority could be different than the WER. For example, for information retrieval purposes, it would be a better choice to prioritize the coverage of domain-specific terms and names over the WER for non-entity words. Thus, we conclude that it depends on the purpose and application what woud be the best choice about including the hotwords functionality in the ASR model or not.}

%\paragraph{Future Work}
We identify three major lines of future work: further finetuning, diarization and better transcription of named entities.

One of \revision{the} major benefits of transformer models from the Huggingface project is that they can easily be finetuned to a downstream task. This is something we tried in this study as well, but the amount of manually transcribed data that could be used as a training set was too small. Yet in future work, with a larger set of transcribed committee meetings, this is an option to improve the output \revision{of} the corpus even more. 

Besides transcriptions, information about who spoke when would also be valuable for political researchers. This can be done with diarization. We tried a zero shot diarization setting with pyannote audio \cite{bredin2020pyannote} yet the results \revision{were} too inaccurate to be useful, and the amount of manually annotated data was too small to train this on our own data. We did not follow up on this pilot yet, as it was outside the scope of this paper. A future study on perfecting the diarization would improve the quality of the corpus for political research. For this a larger training set would be needed.

Another topic that would need further exploration is the recognition of names. 
As mentioned, reliable recognition of named entities would largely increase the number of questions a political scientist could answer using the data. Lines among which this could be achieved would be an even better acoustic model and language model, and maybe also using a more fine grained list of hotwords.

%         ◦ We did x
%         ◦ We found that
%         ◦ Beam search decoding with language model leads to improvement
%         ◦ Hotwords
%         ◦ Huggingface
%         ◦ Already new models on the block with greater potential
%         ◦ Diarization: future work

% \begin{itemize}
%     \item adapting an ASR-pipeline to a specific domain is beneficial.
% \end{itemize}

% \backmatter

% \bmhead{Supplementary information}

% If your article has accompanying supplementary file/s please state so here. 

% Authors reporting data from electrophoretic gels and blots should supply the full unprocessed scans for key as part of their Supplementary information. This may be requested by the editorial team/s if it is missing.

%Please refer to Journal-level guidance for any specific requirements.

\bmhead{Acknowledgments}

%We would like to thank the \href{https://wiki.alice.universiteitleiden.nl/index.php?title=Documentation}{ALICE}\footnote{\url{https://wiki.alice.universiteitleiden.nl/index.php?title=Documentation}} computing cluster for access to the GPU machines and 
We would like to thank the anonymous reviewers for their constructive comments.

% Please refer to Journal-level guidance for any specific requirements.

% \section*{Declarations}

% Some journals require declarations to be submitted in a standardised format. Please check the Instructions for Authors of the journal to which you are submitting to see if you need to complete this section. If yes, your manuscript must contain the following sections under the heading `Declarations':

% \begin{itemize}
% \item Conflict of interest/Competing interests (check journal-specific guidelines for which heading to use)
% \item Ethics approval 
% \item Consent to participate
% \item Consent for publication
% \item Availability of data and materials
% \item Authors' contributions
% \end{itemize}

% \noindent
% If any of the sections are not relevant to your manuscript, please include the heading and write `Not applicable' for that section. 

% %%===================================================%%
% %% For presentation purpose, we have included        %%
% %% \bigskip command. please ignore this.             %%
% %%===================================================%%
% \bigskip

% Springer journals and proceedings: \url{https://www.springer.com/gp/editorial-policies}

% Nature Portfolio journals: \url{https://www.nature.com/nature-research/editorial-policies}

% \textit{Scientific Reports}: \url{https://www.nature.com/srep/journal-policies/editorial-policies}

% BMC journals: \url{https://www.biomedcentral.com/getpublished/editorial-policies}
% \end{flushleft}

\bibliography{sn-bibliography}% common bib file

\newpage
\begin{appendices}

\section{Topics table}\label{appsec:topics}

\begin{table}[h!]
    \centering
    \begin{tabular}{lrlrlrlr}
\toprule
    \multicolumn{2}{l}{TOPIC 1} & \multicolumn{2}{l}{TOPIC 2} & \multicolumn{2}{l}{TOPIC 3} & \multicolumn{2}{l}{TOPIC 4} \\
       word & weight &     word & weight &        word & weight &       word & weight \\
\midrule
      crime & 0.0286 &   people & 0.0295 &       right & 0.0315 &       vote & 0.1052 \\
  terrorism & 0.0207 &     also & 0.0094 &        data & 0.0172 &       open & 0.0842 \\
  organised & 0.0156 &    think & 0.0091 &         law & 0.0153 &  amendment & 0.0583 \\
     victim & 0.0144 &     know & 0.0088 &       state & 0.0151 &     closed & 0.0354 \\
      fight & 0.0112 &     many & 0.0085 &    european & 0.0143 &        one & 0.0352 \\
      human & 0.0102 &    thing & 0.0085 &      member & 0.0127 &      voter & 0.0332 \\
trafficking & 0.0096 &    child & 0.0082 &  commission & 0.0123 &    carried & 0.0326 \\
   criminal & 0.0084 &      one & 0.0082 &  protection & 0.0119 & compromise & 0.0322 \\
   minority & 0.0081 & actually & 0.0079 & fundamental & 0.0118 &   rejected & 0.0227 \\
  hungarian & 0.0079 &   europe & 0.0078 &        rule & 0.0108 &       five & 0.0222 \\
\bottomrule
\end{tabular}

\begin{tabular}{lrlrlrlr}
\toprule
    \multicolumn{2}{l}{TOPIC 5} & \multicolumn{2}{l}{TOPIC 6} & \multicolumn{2}{l}{TOPIC 7} & \multicolumn{2}{l}{TOPIC 8} \\
       word & weight &     word & weight &      word & weight &     word & weight \\
\midrule
       need & 0.0175 &   border & 0.0250 &        mr & 0.0260 &      two & 0.0843 \\
       also & 0.0168 &  country & 0.0180 &     would & 0.0207 & thousand & 0.0439 \\
      state & 0.0166 &   turkey & 0.0177 &     thank & 0.0183 &     year & 0.0334 \\
     member & 0.0148 &     visa & 0.0137 &      like & 0.0161 &  hundred & 0.0283 \\
   european & 0.0124 &   people & 0.0127 & president & 0.0157 &      one & 0.0259 \\
    country & 0.0114 &  refugee & 0.0108 &  question & 0.0148 &   twenty & 0.0233 \\
      think & 0.0113 & schengen & 0.0107 &     think & 0.0146 &     five & 0.0208 \\
     system & 0.0097 &   greece & 0.0099 &      much & 0.0128 &     zero & 0.0166 \\
         eu & 0.0082 &      ngo & 0.0080 &     going & 0.0113 &    three & 0.0162 \\
information & 0.0067 &    italy & 0.0079 & colleague & 0.0108 &  fifteen & 0.0144 \\
\bottomrule
\end{tabular}

    \caption{Table displaying all topics from the topic modeling analysis in section \ref{subsec:topicmodeling}.}
    \label{tab:my_label}
\end{table}

\newpage

\section{Examples of the output of the ASR}\label{secA1}

% \usepackage{multirow}
\begin{table}[h!]
\begin{tabular}{|p{2cm}|p{7cm}|p{1cm}|}
\hline
\textbf{Fragment ID} & \textbf{Reference/Inference}  & \textbf{WER} \\ 
\hline

\multirow{2}{*}{20140723\_068} 
& 
I am glad that the, inside the directive we are strengthening migrants rights, but when the commission talks about the various challenges that exist in implementing the directive, the problem is precisely in the area of workers rights                                                                        
& 
\multirow{2}{*}{0.000} \\ 
\cline{2-2}
& 
i am glad that the inside the directive we are strengthening migrants rights but when the commission talks about the various challenges that exist in implementing the directive the problem is precisely in the area of workers rights                                                                        &
\\ 
\hline

\multirow{2}{*}{20140915\_002} 
& 
yes, I would like to raise a point regarding procedure 
& 
\multirow{2}{*}{0.000} \\ 
\cline{2-2}
& 
yes i would like to raise a point regarding procedure                         
& \\ 
\hline
\multirow{2}{*}{20140701\_020} 
& 
thank you very much colleagues
& 
\multirow{2}{*}{1.000} \\ 
\cline{2-2}
& 
youverymuclege                                                                
\\ 
\hline

\multirow{2}{*}{20140701\_016} 
& 
I would like to propose on behalf of the snd group our colleague claude moraes as chair of this committee, miss zippal, claude moraes our colleague from snd is nominated, any other nominations                                           & 
\multirow{2}{*}{0.394} \\ 
\cline{2-2}
& 
mr i would like to propose on behalf of the sd group our colleague clod morals as the chair of this committee ms zip clodmorarescolleague from sd was nominated any other nominations                                               
& \\ 
\hline
\multirow{2}{*}{20140925\_041} 
& 
state secretaries from Portugal of Hungary, and we had a visit of the president of the European parliament Martin Schultz, all these in the first three days of this week, and this is not an exception, in this context, I am sure it is not a surprise for you that by the end of two thousand thirteen, ninetinine point seven four per cent of the budget had been committed of this agency 
& 
\multirow{2}{*}{0.150} \\ 
\cline{2-2}

& 
the state secretaries from portugal and hungary and we had a visit of the president of the european parliament martin jules all these in the first three days of this week in this is not an exception in this in this context i am sure it is not a surprise for you that by the end of two thousand and thirteen ninety nine seven four of the budget had been committed of this agency       
& \\ 
\hline
\end{tabular} \caption{Examples of the quality of the ASR. This table is discussed in section \ref{subsec:error_analysis}.}\label{tab:outputexamples}
\end{table}

\newpage

\end{appendices}



\end{document}